\documentclass[runningheads]{llncs}
\usepackage{amsmath}
\usepackage{multirow}
\usepackage{tabularx}
\usepackage{booktabs} 
\usepackage{array} 

\usepackage[T1]{fontenc}
\usepackage{graphicx}
\begin{document}
\title{TractShapeNet: Efficient Multi-Shape Learning with 3D Tractography Point Clouds}
%
%
\author{Yui Lo\inst{1,2,4} \and
Yuqian Chen\inst{1,2} \and
Dongnan Liu\inst{4} \and
Jon Haitz Legarreta\inst{1,2} \and
Leo Zekelman\inst{2,7} \and
Fan Zhang\inst{5} \and
Jarrett Rushmore\inst{3,6} \and
Yogesh Rathi\inst{1,2} \and
Nikos Makris\inst{1,3} \and
Alexandra J. Golby\inst{1,2} \and
Weidong Cai\inst{4} \and
Lauren J. O'Donnell\inst{1,2}}

\authorrunning{Y. Lo et al.}
\institute{
Harvard Medical School, Boston, USA \and
Brigham and Women’s Hospital, Boston, USA \and
Massachusetts General Hospital, Boston, USA \and
The University of Sydney, Sydney, Australia \and
University of Electronic Science and Technology of China, Chengdu, China \and
Boston University, Boston, USA \and
Harvard University, Boston, USA \\
\email{odonnell@bwh.harvard.edu}
}
\maketitle              
\begin{abstract}
Brain imaging studies have demonstrated that diffusion MRI tractography geometric shape descriptors can inform the study of the brain’s white matter pathways and their relationship to brain function. In this work, we investigate the possibility of utilizing a deep learning model to compute shape measures of the brain's white matter connections. We introduce a novel framework, TractShapeNet, that leverages a point cloud representation of tractography to compute five shape measures: length, span, volume, total surface area, and irregularity. We assess the performance of the method on a large dataset including 1,065 healthy young adults. Experiments for shape measure computation demonstrate that our proposed TractShapeNet outperforms other point-cloud-based neural network models in both the Pearson correlation coefficient and normalized error metrics. We compare the inference runtime results with the conventional shape computation tool DSI-Studio. Our results demonstrate that a deep learning approach enables faster and more efficient shape-measure computation. We also conduct experiments on two downstream language cognition prediction tasks, showing that shape measures from TractShapeNet perform similarly to those computed by DSI-Studio. Our code will be available at: https://github.com/SlicerDMRI/TractSh-\linebreak apeNet.

\end{abstract}
\section{Introduction}
\label{sec:intro}

Recent studies have shown the potential of tractography shape measures to provide insight into the brain’s structural connections \cite{Yeh2020-ps,Schilling2023-sp} and their relationship to human cognition \cite{Lo2024-cp,Lo2024-zy}. However, existing methods for computation of shape measures can be highly time consuming, particularly when dealing with large-scale diffusion MRI (dMRI) tractography datasets including thousands of participants and many millions of tractography streamlines. One challenge is that existing methods require an intermediate step to convert geometric tractography streamline data to an image data representation using a voxel grid \cite{Yeh2020-ps,Schilling2023-sp}. Computing shape measures directly from geometric tractography data, while reducing runtime and enhancing computational efficiency, remains a challenge. 

\begin{figure}[htbp]
\centering
\includegraphics[width=\linewidth]{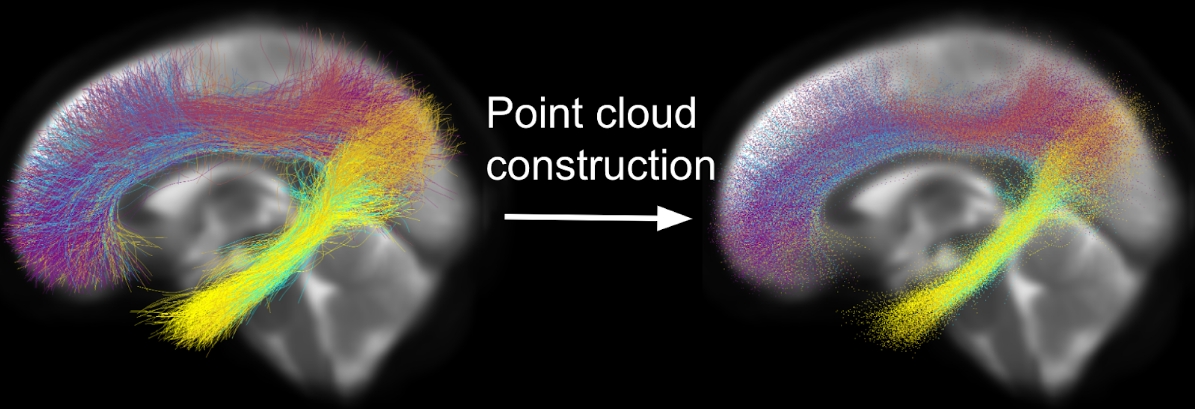}
\caption{An example white matter connection, the cingulum bundle (CB, top image), extracted from the entire white matter of the human brain using an atlas-based fiber clustering approach \cite{Zhang2018-ez}. The streamlines of the 22 fiber clusters in the CB are converted to point cloud representations (bottom).} \label{fig1}
\end{figure}

In recent years, deep learning methods have been successful in dMRI tractography analysis \cite{Karimi2024-wy}, but computing shape measures in a way that can best take advantage of deep networks remains an open challenge. Deep learning using point clouds is promising for the analysis of geometric structures in fields like 3D object retrieval and medical imaging \cite{Guo2021-yw}. A point cloud is a finite set of points with 3D coordinates and optional attributes such as normals or features. In dMRI tractography, white matter connections are represented as streamlines that consist of sequences of points (Fig. 1). Recent works have demonstrated the potential of point-based neural networks for the analysis of dMRI tractography in tasks such as parcellation, filtering, and prediction of cognitive performance \cite{Xue2023-ju,Chen2024-fh,Astolfi2020-qp}.

To the best of our knowledge, no deep learning methods have focused on computing white matter tractography shape, and point-cloud-based deep networks have not yet been applied to tractography to compute relevant shape features. We identify a research gap where efficient deep learning models have the potential to improve the runtime and processing speed, making it feasible to deploy shape analyses on very large brain tractography datasets. 

The main contribution of this work is to explore the potential of a deep point cloud model to compute a set of tractography shape measures. We propose an end-to-end, time-efficient, multi-shape learning strategy to compute individual shape measures using individual fiber clusters as input as shown in Fig. 2.

\begin{figure}[htbp]
\centering
\includegraphics[width=\columnwidth]{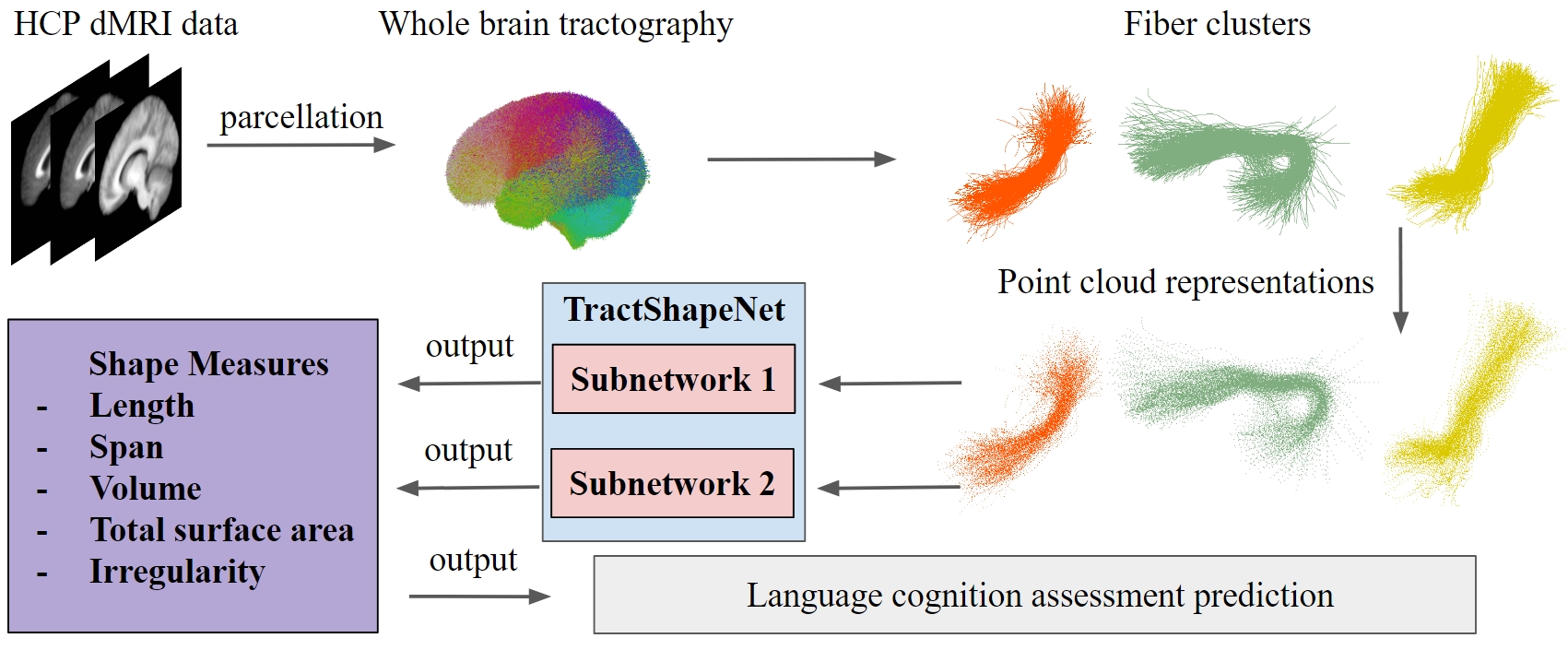}
\caption{Overview of TractShapeNet.}
\label{fig2}
\end{figure}

\section{Methods}
\label{sec:methods}
\subsection{dMRI Dataset and Tractography Fiber Clustering}
This study evaluates our proposed method on the large-scale Human Connectome Project Young Adult (HCP-YA) dataset, which includes 1,065 healthy young adults (575 females and 490 males, 28.7 years old on average) \cite{Van_Essen2013-yo,Van_Essen2012-qh}. Whole brain tractography is generated from each subject’s dMRI data using a two-tensor unscented Kalman filter method \cite{Malcolm2010-fj}. Tractography is then parcellated into individual fiber clusters using an anatomically curated tractography brain atlas \cite{Zhang2018-ez}. We focus on the key task of shape computation for association pathways \cite{Yeh2020-ps}. We employ fiber clusters within ten left hemisphere association tracts of arcuate fasciculus (AF), cingulum bundle (CB), extreme capsule (EmC), inferior longitudinal fasciculus (ILF), middle longitudinal fasciculus (MdLF), and uncinate fasciculus (UF). This results in 73 fiber clusters per subject, for a total of 77,745 clusters across the 1,065 subjects for training and testing our methods.

\subsection{Shape Measures}
We compute ground-truth shape measures for each fiber cluster using DSI-Studio \cite{Yeh2020-ps}. Detailed definitions of the shape measures can be seen in \cite{Yeh2020-ps,Lo2024-zy}. We select five shape descriptors for analysis in this study: length, span, volume, total surface area, and irregularity. Length is the average streamline length of the fiber cluster, span is the distance between the two ends of the fiber cluster, and volume is the volume of voxels occupied by the fiber cluster. We include total surface area and irregularity because we have shown that they are highly informative for the prediction of individual cognitive performance \cite{Lo2024-zy}. Total surface area is the area of the outermost or surface voxels occupied by the fiber cluster, while irregularity quantifies how different the fiber cluster shape is from an idealized cylinder.

\subsection{Representing Fiber Clusters as Point Clouds}
Our model is trained using the right-anterior-superior (RAS) coordinate points of individual fiber clusters as independent (input) variables, with the shape measures as the dependent (output) variables. Following TractGeoNet \cite{Chen2024-fh}, we represent each fiber cluster as a point cloud. We construct 73 point clouds for every subject, corresponding to the 73 fiber clusters. We leverage a random sampling strategy to select $N$ random points from each cluster as input for the deep learning model. Random sampling has previously been shown to be an effective strategy for the analysis of tractography as point clouds \cite{Chen2024-fh} because this strategy achieves an equal probability of selection across all points \cite{Peters1995-qn}.

\subsection{Network Architecture and Proposed Loss Function}
This work proposes a novel multi-shape learning framework that aims to efficiently output shape measurements of the fiber clusters with a deep learning approach. The overall pipeline of the proposed work is shown in Fig. 2. The input to the model is the point cloud representation of fiber clusters constructed in Section 2.3. We follow a Siamese network design containing two identical subnetworks with shared weights \cite{Bromley1993-bl,Chen2024-fh}. The identical subnetworks adopt the PointNet architecture, which has been widely applied for processing point clouds \cite{Aoki2019-vk,Charles2017-gr}, including tractography data \cite{Xue2023-ju,Chen2024-fh}. We retain the MLP network design and remove the spatial transformation net for our subnetwork architecture to preserve fiber tract anatomical size information \cite{Xue2023-ju,Chen2024-fh} and to reduce floating-point operations per second for better efficiency \cite{Aoki2019-vk}. 

We introduce a hybrid loss function $L_{total}$ that combines the subnetwork loss functions ($L_1$, $L_2$) with a proposed weighted pairwise Siamese-Fourier loss function ($L_{SF}$) to optimize the model performance as shown in Equation (1). The subnetwork loss functions calculate the error between the output and the ground-truth. The weighted pairwise loss function $L_{SF}$ complements this by capturing patterns across a Fourier frequency domain. As shown in Equation (2), we utilize the discrete Fourier transformation (DFT) to transform our predicted output and ground-truth shape measures into Fourier frequency domain representations \cite{Cooley1965-co,Park2022-mo}. This transform can potentially enhance multi-shape learning by leveraging global relationships across the different shape measures. $e^{-\tfrac{j2\pi }{N}kn}$ is Euler's formula that transforms the Fourier frequency domain, where $k$ and $n$ represent the $k$th frequency component and the $n$th sample, respectively. $L_{SF}$ calculates the mean squared error between the output differences ($O_1-O_2$) and corresponding ground-truth differences ($GT_1-GT_2$) in Fourier frequency space: 

\begin{equation}
L_{\text{total}} = L_1 + L_2 + \alpha \times L_{SF},
\end{equation}

\begin{equation}
L_{SF} = \frac{1}{N} \sum_{k=0}^{N-1} \Bigg( \left| \sum_{n=0}^{N-1} (O_1 - O_2) \cdot e^{-\frac{j 2 \pi}{N} k n} \right| - \left| \sum_{k=0}^{N-1} (GT_1 - GT_2) \cdot e^{-\frac{j 2 \pi}{N} k n} \right| \Bigg)^2.
\end{equation}

\subsection{Implementation Details}
The model is trained and evaluated with a train test split of 80\% and 20\%. The model is set with a batch size of 128 for 200 epochs. The initial learning rate is 0.0005 with the Adam optimizer \cite{Kingma2014-lm} with a weight decay of 0.005. The scheduler updates the learning rate with a decay factor (gamma) of 0.1 every 200 steps. We tuned the weight ($\alpha$) of the Siamese Fourier loss function and set it to 3 for optimal performance. The code is implemented using PyTorch v1.13 \cite{Paszke2019-nb}. All experiments are conducted on the Jetstream2 cloud computing environment, configured with an NVIDIA A100 40 GB GPU, 117 GB of RAM, and 32 CPU cores \cite{Hancock2021-ro,Boerner2023-bo}.

\section{Experiments and Results}
\subsection{Evaluation Metrics}
We employ the Pearson correlation coefficient (Pearson’s $r$) \cite{Sedgwick2012-ur} and the normalized mean squared error (nMSE) to evaluate model performance. Pearson’s $r$ measures the strength and direction (positive or negative) of the correlation between predicted and ground-truth shape measurements. nMSE measures the error between the predicted and ground-truth values \cite{Li2021-bu,Lemkaddem2014-qv}. nMSE is normalized to a range between 0 and 1, where an nMSE value close to 0 represents a low error when compared to the ground-truth.

\subsection{Evaluation of Model Performance}
We compared our proposed TractShapeNet with different baseline methods, PointNet and TractGeoNet, as shown in Tables 1 and 2. PointNet is a state-of-the-art baseline method for deep learning utilizing point clouds \cite{Charles2017-gr}. TractGeoNet is a state-of-the-art network for performing regression using tractography data represented as point clouds \cite{Chen2024-fh}. We trained these networks on our shape prediction task, including modification of their output layers and fine tuning, with implementation details as described in Section 2.5.

\begin{table}[htbp]
\centering
\caption{Comparison of methods using Pearson correlation coefficient ($r$). The best result in each row is in bold.}
\label{tab1}
\begin{tabularx}{\columnwidth}{|>{\centering\arraybackslash}X|>{\centering\arraybackslash}X|>{\centering\arraybackslash}X|>{\centering\arraybackslash}X|}
\hline
\textbf{Shapes} & \textbf{PointNet} \cite{Charles2017-gr} & \textbf{TractGeoNet} \cite{Chen2024-fh} & \textbf{TractShapeNet} \\
\hline
Length &0.970 &0.975 &\textbf{0.985}\\\hline

Span &
0.977 &
0.980 &
\textbf{0.987} \\\hline

Volume &
0.710 &
0.722 &
\textbf{0.766} \\\hline

Total Surface Area &
0.827 &
0.836 &
\textbf{0.872} \\\hline

Irregularity &
0.766 &
0.796 & 
\textbf{0.867}\\
\hline
Average & 0.850 ± 0.120 & 0.862  ± 0.113 & \textbf{0.895 ± 0.093}\\
\hline
\end{tabularx}
\end{table}

\begin{table}[htbp]
\centering
\caption{Comparison of methods using nMSE. The best result in each row is in bold.}
\label{tab2}
\begin{tabularx}{\columnwidth}{|>{\centering\arraybackslash}X|>{\centering\arraybackslash}X|>{\centering\arraybackslash}X|>{\centering\arraybackslash}X|}
\hline
\textbf{Shapes} & \textbf{PointNet} \cite{Charles2017-gr} & \textbf{TractGeoNet} \cite{Chen2024-fh} & \textbf{TractShapeNet} \\
\hline
Length &0.065&
0.054&
\textbf{0.031}
\\\hline

Span &0.052
 &0.044
 &
\textbf{0.027} \\\hline

Volume & 0.503
 & 0.485
 &
\textbf{0.419} \\\hline

Total Surface Area & 0.324
 & 0.309
 & 
\textbf{0.245} \\\hline

Irregularity & 0.439
 & 0.381
 & 
\textbf{0.253}\\
\hline
Average &0.277 ± 0.209 & 0.254 ± 0.198 & \textbf{0.195 ± 0.167}\\
\hline
\end{tabularx}
\end{table}

Tables 1 and 2 give the Pearson correlation coefficient ($r$) and the nMSE performance for the compared methods. The results show that point-based deep learning enables computation of shape measures. We observe that TractShapeNet has the highest performance across all shape measures. This is apparent particularly in the more challenging shape measures known to have higher anatomical variability or lower reliability, such as volume, total surface area, and irregularity \cite{Yeh2020-ps}. This suggests that the proposed TractShapeNet is better at learning these shape features from input point clouds.

\subsection{Evaluation of Inference Runtime}
We calculate the average inference runtime in milliseconds (ms) to compute the five shape measures for individual fiber clusters. Results in Table 3 demonstrate that deep point cloud models show faster inference runtimes than DSI-Studio. Although TractShapeNet is slightly less efficient than TractGeoNet, it offers a tradeoff with improved performance across all shape measures. 

\begin{table}[htb]
\centering
\caption{Comparisons of inference runtime in milliseconds across methods. The fastest inference runtime is in bold.}
\label{tab3}
\begin{tabularx}{\columnwidth}{|>{\centering\arraybackslash}X|>{\centering\arraybackslash}X|>{\centering\arraybackslash}X|>{\centering\arraybackslash}X|>{\centering\arraybackslash}X|}
\hline
\textbf{Methods} & \textbf{DSI-Studio} \cite{Yeh2020-ps} & \textbf{PointNet} \cite{Charles2017-gr} & \textbf{TractGeoNet} \cite{Chen2024-fh} & \textbf{TractShape-Net} \\
\hline
Runtime & 41.10 ±21.3987 &
1.44 ± 0.0003 ms &
\textbf{0.89 ± 0.0001ms}&
1.24 ± 0.0004ms
\\
\hline
\end{tabularx}
\end{table}

\subsection{Evaluation of Downstream Testbed Task}

To investigate the effectiveness of shape measures from TractShapeNet, we perform a widely applied downstream testbed task, the prediction of subject-specific language performance on the NIH Toolbox Picture Vocabulary Test (TPVT) and NIH Toolbox Oral Reading Recognition Test (TORRT) \cite{Lo2024-cp,Chen2019-hf,Gershon2013-eu,Weintraub2013-te,Lo2024-zy}. Using subject-specific shape measures as inputs, we leverage a least absolute shrinkage and selection operator (LASSO) machine learning model to predict TPVT and TORRT \cite{Tibshirani2018-os,Lo2024-zy}. Results in Table 4 show that the shape measures from TractShapeNet perform similarly to those computed by DSI-Studio. We believe the performance on these downstream tasks illustrates that TractShapeNet has captured some intrinsic white matter tractography shape information. This comparison demonstrates that deep point cloud models can generate informative shape measures for studying the brain’s white matter and its relationship to language cognition.

\begin{table}[htb]
\centering
\caption{Prediction Performance for TPVT and TORRT ($r$). Best performances are in bold.}
\label{tab4}
\begin{tabularx}{\columnwidth}{|>{\centering\arraybackslash}X|>{\centering\arraybackslash}X|>{\centering\arraybackslash}X|>{\centering\arraybackslash}X|>{\centering\arraybackslash}X|}
\hline
\multirow{2}{*}{} & \multicolumn{2}{c|}{\textbf{TPVT}} & \multicolumn{2}{c|}{\textbf{TORRT}} \\
\cline{2-5}
\textbf{Shapes} &\textbf{DSI-Studio} \cite{Yeh2020-ps} & \textbf{TractShape- Net} & \textbf{DSI-Studio} \cite{Yeh2020-ps} & \textbf{TractShape- Net} \\
\hline
Length & 0.195 & \textbf{0.230} & 0.096 & \textbf{0.184} \\
\hline
Span & \textbf{0.176} & 0.170 & \textbf{0.220} &0.123 \\
\hline
Volume & 0.152 & \textbf{0.164} & \textbf{0.182} & 0.160 \\
\hline
Total Surface Area & \textbf{0.191} & 0.131 & 0.185 & \textbf{0.214} \\
\hline
Irregularity & 0.165 & \textbf{0.211} & 0.131 & \textbf{0.197} \\
\hline
Average & 0.176 ± 0.018 & \textbf{0.181 ± 0.039} & 0.163 ± 0.049 & \textbf{0.175 ± 0.035} \\
\hline
\end{tabularx}
\end{table}

\section{Conclusion}
In this study, we proposed a novel deep learning framework to compute tractography shape measures with successful application on the HCP-YA dataset. Our proposed model outperforms other point cloud models across all shape measures. Our proposed model enables efficient and effective fiber cluster shape computation.

\section{Compliance with ethical standards}
\label{sec:ethics}
This study uses public HCP imaging data; no ethical approval was required.

\section{Acknowledgements} 
We gratefully acknowledge funding provided by the following grants: National Institutes of Health (NIH) grants R01MH132610, R01MH125860, R01MH119222, R01NS12-5307, R01NS125781, and R21NS136960. FZ is in part supported by National Key R\&D Program of China (No. 2023YFE0118600) and the National Natural Science Foundation of China (No. 62371107). This work is also supported by The University of Sydney International Scholarship and Postgraduate Research Support Scheme.

%
%
%
\bibliographystyle{splncs04}
\bibliography{reference.bib}
\end{document}